\pdfoutput=1

\documentclass[journal]{IEEEtran}

\usepackage{cite}
\usepackage{amsmath,amssymb,amsfonts}
\usepackage{algorithm}
\usepackage{algorithmic}
\usepackage{graphicx}
\usepackage[hidelinks]{hyperref}
\usepackage{booktabs}
\usepackage{multirow}
\usepackage{arydshln}
\usepackage{etoolbox}
\def\BibTeX{{\rm B\kern-.05em{\sc i\kern-.025em b}\kern-.08em
    T\kern-.1667em\lower.7ex\hbox{E}\kern-.125emX}}
\providecommand{\refname}{References}

\makeatletter
\patchcmd{\thebibliography}{\itemsep 0pt plus pt\relax}{\itemsep 0pt plus 1pt\relax}{}{}
\makeatother
\begin{document}
\title{HyDAR-Pano3D: A Hybrid Disentangled Anatomical Recovery Framework for Panoramic-to-3D Reconstruction}
\author{Yaoyao YUE, Jérôme Schmid, Xiaoshuang Li, Eduardo Delamare, and Jinman KIM, \IEEEmembership{Member, IEEE}
\thanks{Yaoyao Yue and Jinman Kim are with the School of Computer Science, the University of Sydney, Australia (e-mail: yyue5992@uni.sydney.edu.au; jinman.kim@sydney.edu.au).}
\thanks{Jérôme Schmid is with the Geneva School of Health Sciences, HES-SO University of Applied Sciences and Arts Western Switzerland, Geneva, Switzerland (e-mail: jerome.schmid@hesge.ch).}
\thanks{Xiaoshuang Li is with Department of Computer Science and Engineering, Shanghai Jiao Tong University, Shanghai 200240, China  (e-mail: lixiaoshuang@sjtu.edu.cn).}
\thanks{Eduardo Delamare is with Sydney Dental School, Faculty of Medicine and Health, The University of Sydney, Sydney, Australia (e-mail: eduardo.delamare@sydney.edu.au).}}

\maketitle

\begin{abstract}
Panoramic radiograph (PR) is fundamentally used in routine dental care, but it inherently provides only a two-dimensional (2D) projection of complex three-dimensional (3D) craniofacial anatomy. Most existing learning-based methods attempt to computationally recover this 3D information by directly regressing native cone-beam computed tomography (CBCT) volumes from PR. However, this direct mapping requires the model to simultaneously learn common anatomical structures and patient-specific morphological variations. This entangled formulation makes the ill-posed 2D-to-3D inverse problem highly ambiguous, often producing over-smoothed reconstructions with blurred anatomical boundaries. To address this, we propose HyDAR-Pano3D, a two-stage framework that reformulates PR-to-CBCT reconstruction as a disentangled anatomical recovery problem. In Stage 1, a dual-encoder network integrates radiographic features with SAM-derived semantic priors to reconstruct an arch-normalized canonical volume. In Stage 2, an Anatomical Restoration Network predicts a prior-constrained structured deformation field to map this canonical volume back to the native space, restoring individual morphological variations. Experiments on three large-scale datasets show that HyDAR-Pano3D significantly outperforms baseline methods ($p < 0.05$), achieving a 25.76 dB PSNR, 85.70\% SSIM, and an 83.83\% overall anatomical Dice score. The synthesized volumes successfully support downstream segmentation of whole teeth (82.4\% Dice) and the inferior alveolar canal (72.2\% Dice), demonstrating that our disentangled approach preserves clinically relevant structures to enable robust anatomy-aware assessment when CBCT data is unavailable.
\end{abstract}


\section{Introduction}
\label{sec:introduction}
\IEEEPARstart{P}{anoramic} radiograph (PR) is a fundamental dental imaging modality for diagnosis and treatment planning. PR provides a comprehensive overview of the upper and lower jaws with low radiation dose, low cost, and high acquisition efficiency \cite{ludlow_patient_2008, mallya_white_2019, izzetti_basic_2021}. Due to these advantages, PR is widely adopted for initial evaluation in orthodontics, implant planning, and maxillofacial screening. However, PR is inherently a two-dimensional (2D) projection of complex three-dimensional (3D) anatomy acquired along a curved dental arch trajectory. This imaging mechanism introduces non-linear geometric distortion, depth compression, and severe anatomical superposition, particularly in the buccolingual direction \cite{suomalainen_dentomaxillofacial_2015}. As a result, substantial spatial information is irreversibly lost during projection, limiting the reliability of PR for volumetric analysis and precise anatomical localization \cite{antony_two-dimensional_nodate, tandogdu_comparison_2022, lopes_comparison_nodate}.

Cone-beam computed tomography (CBCT) alleviates these limitations by providing isotropic 3D volumetric representations of dental anatomy, allowing an accurate assessment of tooth morphology, alveolar bone structure, and critical neurovascular anatomy \cite{leung_accuracy_2010, scarfe_maxillofacial_2012}. Nevertheless, CBCT acquisition is associated with higher radiation exposure and increased cost compared to conventional radiograph \cite{thongvigitmanee_radiation_2013}. According to the ALARA principle, clinical guidelines recommend CBCT only when 2D imaging is insufficient to support diagnostic decisions \cite{jaju_cone-beam_2015}. This clinical trade-off motivates the development of computational approaches capable of estimating patient-specific 3D anatomical information from a single panoramic radiograph, thus extending the clinical utility of PR without additional radiation burden.

Recent advances in deep learning have demonstrated the feasibility of 2D-to-3D reconstruction from PR~\cite{song_oral-3d_2021, linguraru_px2tooth_2024, parida_vit-nebla_2025, mohamed_3d_2025}. However, most existing methods formulate this task as a direct mapping from a single 2D projection to a full 3D volume in the native anatomical space. This direct regression approach is fundamentally limited because it forces the network to simultaneously overcome two distinct and severe challenges. First, the network must hallucinate missing depth cues and resolve severe projection overlap (information loss). Second, it must account for massive patient-specific morphological variations in dental arch geometry caused by the non-linear PR acquisition process~\cite{suomalainen_dentomaxillofacial_2015, scarfe_what_2008}.

This entangled formulation—attempting to jointly model fine-grained volumetric appearance and large-scale, arch-dependent spatial deformation—makes the inverse mapping highly ambiguous. As a consequence, the variance of the target distribution increases significantly, causing these networks to produce population-averaged, over-smoothed reconstructions with blurred anatomical boundaries and loss of critical structural details.

To address this core contradiction, we propose HyDAR-Pano3D, a novel framework that reformulates PR-based 3D reconstruction from an ill-posed direct mapping into a disentangled anatomical recovery problem. Our core insight is to decouple the recovery of high-fidelity volumetric density from the large-scale spatial deformation introduced by individualized arch geometry.

We realize this divide and conquer strategy through a highly constrained two-stage pipeline. In the first stage, rather than struggling with patient-specific geometric variance, we focus entirely on resolving depth ambiguity by synthesizing the 3D volume in a standardized, arch-normalized canonical space. To achieve this, we introduce a dual-encoder architecture with an Adaptive Cross-attention Fusion (ACF) mechanism. This allows us to inject high-level semantic priors from domain-adapted vision foundation models directly into the radiographic feature space, effectively recovering the fine structural details lost to 2D projection overlap. In the second stage, with the high-fidelity canonical anatomy established, an Anatomical Restoration Network (AR-Net) utilizes a prior-guided structured deformation field to warp the canonical volume back into the patient's native CBCT space, thus restoring individualized morphological variation.

By explicitly separating prior-guided anatomy synthesis from individualized morphology restoration, HyDAR-Pano3D drastically reduces the cognitive load on the network, enabling high-fidelity, patient-specific 3D reconstructions.

The main contributions of this work are as follows:
\begin{itemize}
    \item A Disentangled Paradigm for PR-to-CBCT Inversion: We formulate the severely ill-posed PR reconstruction task by decoupling volumetric density estimation from arch-dependent geometric distortion. Our two-stage approach significantly constrains the inverse mapping: resolving complex density inference within an arch-normalized space, while capturing patient-specific morphology through robust geometric deformation.
    
    \item Bridging 2D Semantic Priors to 3D Volumetric Recovery: To overcome the inherent depth ambiguity and information loss of a single 2D projection, we introduce a dual-encoder architecture with an Adaptive Cross-attention Fusion (ACF) mechanism. This design propagates robust structural priors from vision foundation models to guide the recovery of fine anatomical details.
    
    \item Unprecedented Anatomical Fidelity for Downstream Utility:  We extensively validate HyDAR-Pano3D across three large-scale datasets, demonstrating state-of-the-art performance. The synthesized volumes successfully drive critical downstream clinical tasks—achieving high accuracy in both whole-tooth and inferior alveolar canal (IAC) segmentation—proving the framework's viability as a low-dose surrogate for anatomy-aware clinical screening.
\end{itemize}

\section{Related Work}
\subsection{Cross-Modality 2D-to-3D Reconstruction}
Reconstructing 3D volumetric data from limited 2D projections is a long-standing problem in dental medical imaging, motivated by the need to reduce radiation exposure and acquisition costs.
Early approaches addressed this linear inverse problems using analytical methods such as filtered back-projection, which typically require multiple angular views.
With the advent of deep learning, data-driven methods have been developed to tackle highly ill-posed sparse-view reconstruction problems.
Ying et al.~\cite{ying_x2ct-gan_2019} proposed X2CT-GAN to synthesize CT volumes from biplanar PR using generative adversarial networks.
More recently, implicit neural representations, including Neural Radiance Fields, have been adapted to medical imaging (e.g., MedNeRF~\cite{corona-figueroa_mednerf_2022, avidan_conditional-flow_2022}), enabling continuous volumetric estimation from sparse projections.

However, most general-purpose X-ray-to-CT frameworks assume linear perspective projection geometry or rely on multi-view consistency constraints, such as orthogonal or biplanar acquisitions.
These assumptions break down in PR, where a single projection is acquired along a non-linear, curved tomographic trajectory~\cite{mallya_white_2019}.
As a result, conventional cross-modality reconstruction paradigms fail to model the focal trough distortion and complex anatomical superposition inherent to panoramic imaging, necessitating specialized reconstruction strategies.

\begin{figure*}[t]
    \centering
    \includegraphics[width=0.95\linewidth,height=7.5in,keepaspectratio]{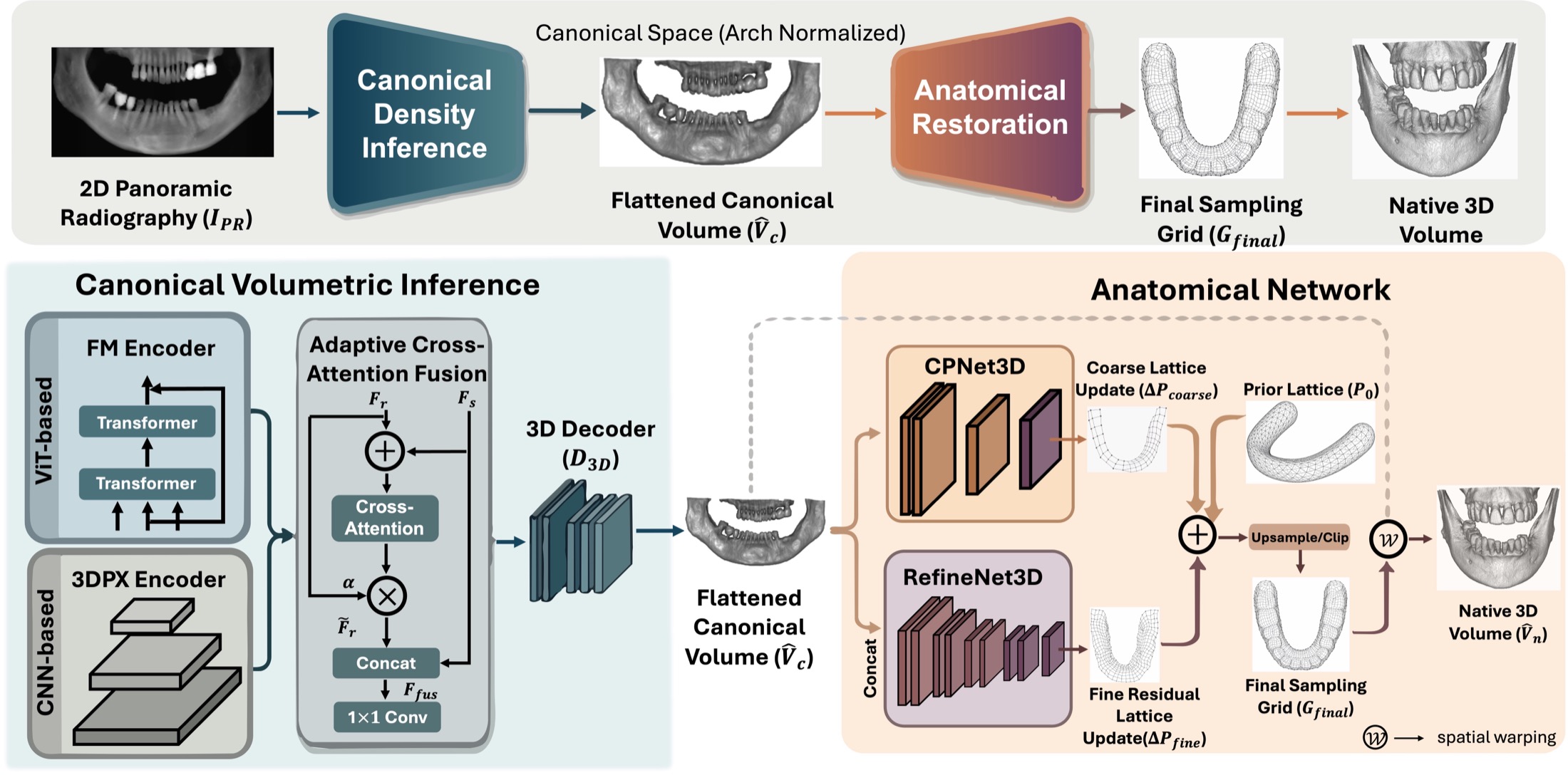}
    \caption{HyDAR-Pano3D Overview. The framework factorizes 3D dental reconstruction into: (1) Canonical Volumetric Inference: Stage 1 maps 2D panoramic space to an arch-normalized 3D canonical volume ($\hat{V}_c$) via dual-stream encoders (3DPX, SemiT-SAM) and adaptive cross-attention (ACF). (2) Anatomical Restoration (AR-Net): Stage 2 restores patient-specific morphology via hierarchical control-lattice deformation. Coarse ($\Delta P_{\mathrm{coarse}}$) and fine residual ($\Delta P_{\mathrm{fine}}$) updates are integrated with an arch-shaped prior ($P_0$) to generate a sampling grid ($G_{\mathrm{final}}$) for warping $\hat{V}_c$ into the native reconstruction ($\hat{V}_n$).}
    \label{fig:pipeline}
\end{figure*}

\subsection{Panoramic-to-3D Dental Reconstruction}
Recent studies in dental imaging have explored reconstructing 3D anatomical structures directly from single panoramic radiographs.
Existing approaches can be broadly categorized into geometry-centric and volume-centric frameworks.

\subsubsection{Geometry-Centric Approaches}
Geometry-centric methods focus on recovering surface representations of specific anatomical structures, namely teeth and mandible. DTR-Net~\cite{mei_dtr-net_2024} introduced a dual-space formulation to reconstruct tooth meshes by leveraging statistical shape priors. Similarly, X2Teeth~\cite{linguraru_px2tooth_2024} and PX2Tooth~\cite{martel_x2teeth_2020} employ segmentation-driven or point-cloud-based networks to infer 3D dental morphology. To address the dental arch geometry, X2Teeth fits a parametric dental-arch model and bends the reconstructed teeth to align with the estimated arch, whereas PX2Tooth restores tooth poses through a global 3D registration that arranges individually predicted teeth into a coherent dental arch. 

While effective for applications emphasizing surface topology (e.g., orthodontic analysis), these methods are typically constrained by predefined shape models and thus have limited capacity to capture topological variations such as fractures or fragmented bone structures. More fundamentally, geometry-centric approaches operate on surface representations and therefore cannot recover internal volumetric density, such as trabecular bone structure, which is essential for radiometric assessment. Consequently, they are insufficient for diagnostic tasks that require volumetric information, including the detection of periapical lesions or evaluation of bone quality for implant planning.

\subsubsection{Volume-Centric Approaches}
Volume-centric methods aim to reconstruct the full volumetric density field (pseudo-CBCT) directly in the native anatomical coordinate system. Oral-3D~\cite{song_oral-3d_2021} proposed a two-stage framework that incorporates projection-physics constraints to recover partial oral volumes. More recently, Park et al.~introduced NeBLa~\cite{park_3d_2023} and its successor ViT-NeBLa~\cite{parida_vit-nebla_2025}, which employ differentiable projection operators based on the Neural Beer--Lambert Law to invert panoramic radiographs. These methods represent the current state of the art in volumetric panoramic reconstruction. However, they generally formulate the task as a direct mapping from a panoramic image $I_{\text{PR}}$ to a volumetric representation $V_{\text{CBCT}}$ in the native CBCT space. Under this formulation, the network must simultaneously account for arch-dependent 2D appearance and patient-specific 3D anatomical curvature, which increases the complexity and variability of the inverse mapping. This challenge motivates a reconstruction framework that explicitly disentangles anatomical geometry from volumetric density.

\subsubsection{Canonical Spaces and Deformable Registration}
Separating shape variability from intensity variability is a well-established concept in computational anatomy. Atlas-based analysis and deformable registration frameworks, such as VoxelMorph~\cite{balakrishnan_voxelmorph_2019}, typically map patient data into a common reference space to facilitate comparison and learning. In volumetric reconstruction, learning in a canonical space enables standardization of anatomical pose and reduces inter-subject geometric variance. While canonical-space representations are widely used in brain MRI analysis and cardiac modeling, their application to the panoramic inverse problem remains underexplored. To our knowledge, existing panoramic reconstruction methods do not explicitly decouple anatomical density inference from patient-specific geometric deformation. By decomposing the task into canonical volume synthesis followed by deformation field estimation, the optimization landscape can be substantially regularized compared with direct native-space inference. In this work, we explicitly adopt this paradigm, utilizing a canonical space to normalize arch variations before predicting patient-specific deformations.

\subsubsection{Semantic Priors for Ill-Posed Reconstruction}
While mapping to a canonical space reduces arch-dependent geometric variance, recovering 3D volumetric density from a single 2D projection remains ill-posed due to inherent depth ambiguity and extreme anatomical superimposition. To effectively invert such information loss, reconstruction frameworks must rely on strong structural priors. Extensive literature in computational imaging has demonstrated that learned semantic priors can significantly stabilize inverse problems and bias solutions toward anatomically plausible configurations \cite{ongie_deep_2020, antun_instabilities_2020, huang_considering_2021}.

Recently, large-scale vision foundation models have emerged as powerful extractors of these robust semantic and structural priors. Models such as the Segment Anything Model (SAM) \cite{kirillov_segment_2023} and its domain-adapted variants (e.g., MedSAM \cite{ma_segment_2024}, SemiT-SAM \cite{wang_semit-sam_2025}) excel at capturing complex morphological boundaries and high-level anatomical topology. While initially designed for 2D segmentation tasks, recent evidence suggests that dense semantic embeddings from these foundation encoders can provide crucial guidance for medical image synthesis and reconstruction \cite{morshuis_segmentation-guided_2024, feng_utility_nodate}.

However, naively applying these 2D models to 3D reconstruction is non-trivial due to the modality gap between dense volumetric fields and 2D radiographic features. In this work, we bridge this gap by extracting high-level semantic priors from domain-adapted foundation models and injecting them into the canonical volume synthesis process. This explicitly guides the network to recover fine-grained dental details that are otherwise lost to projection overlap, functioning as a powerful regularizer for the disentangled inverse mapping.

\section{Methodology}
\subsection{Overview}

We propose HyDAR-Pano3D, a two-stage framework to recover patient-specific 3D dental anatomy from a single PR. To overcome the ill-posed nature of 2D-to-3D inversion, our core insight is the explicit disentanglement of volumetric inference from individualized geometric restoration. 

As illustrated in Fig.~\ref{fig:pipeline},  HyDAR-Pano3D introduces an intermediate canonical space that normalizes global dental arch curvature, significantly reducing inter-subject geometric variability. The overall factorization is written as:

\begin{equation}
    \hat{V}_{c} = f_{\theta}{I_{\mathrm{PR}}}, \qquad
    \hat{P} =  g_{\psi}{\hat{V}_{c}}, \qquad
    \hat{V}_{n} = \mathcal{W}(\hat{V}_{c}, \hat{P}),
\end{equation}

where $I_{\mathrm{PR}}$ denotes the input PR, $\hat{V}_{c}$ denotes the reconstructed flattened canonical volume, $\hat{P}$ denotes the predicted deformation parameters, $\mathcal{W}$ denotes the spatial  warping operator, and $\hat{V}_{n}$ denotes the native-space reconstruction.

Stage~1 learns a canonical volume representation that preserves local anatomical intensity patterns while suppressing patient-specific arch geometry. Stage~2 restores subject-specific global and local geometry by predicting a lattice-based deformation that maps the canonical volume back to the native anatomical space. Unlike direct PR-to-CBCT pipelines that jointly regress density and geometry in a single step, this formulation explicitly separates canonical density synthesis from geometric restoration, yielding a better-constrained inverse mapping. The methodology is structured to reflect our core contributions: Section~\ref{sec:stage1} details the foundation-model-guided synthesis in the canonical space, resolving 2D-to-3D depth ambiguity; Section~\ref{sec:stage2} introduces AR-Net for patient-specific anatomical restoration; and the final part of this section outlines the training objectives.

\subsection{Stage~1: Canonical Volumetric Synthesis via Foundation-Model Guidance}
\label{sec:stage1}
The primary objective of Stage 1 is to synthesize a high-fidelity 3D volume in an arch-normalized canonical space. By explicitly stripping away the large-scale, patient-specific arch curvature, this stage focuses exclusively on recovering consistent localized anatomy (e.g., teeth, roots, and surrounding alveolar bone).

\textbf{Dual-stream encoding for mitigating depth ambiguity.}
Even in a canonical space, a single PR exhibits severe depth ambiguity due to anatomical superposition. To alleviate this and hallucinate missing depth cues, we propose a dual-stream encoder. This design extracts complementary radiographic appearance cues and high-level structural priors, which are subsequently integrated to facilitate depth disambiguation in overlapping regions.

For the radiographic stream, we employ our previously validated 3DPX encoder~\cite{linguraru_3dpx_2024} to extract local texture and spatial contextual features. At the $l$-th scale, the spatial feature map is denoted as $F_r^{(l)} = E_r^{(l)}(I_{\mathrm{PR}})$, where $I_{\mathrm{PR}} \in \mathbb{R}^{1 \times H \times W}$. In parallel, to extract robust geometric boundaries, we introduce a semantic stream instantiated with SemiT-SAM~\cite{wang_semit-sam_2025}, a Vision Transformer (ViT) explicitly domain-adapted for PR. We map the single-channel radiograph to a 3-channel tensor via channel replication, $I_{\mathrm{PR}}^{(3)} = \mathrm{Rep3}(I_{\mathrm{PR}}) \in \mathbb{R}^{3 \times H \times W}$, yielding the multi-scale semantic spatial representations $F_s^{(l)} = E_s^{(l)}(I_{\mathrm{PR}}^{(3)})$. 

\textbf{Multi-scale Adaptive Cross-Attention Fusion (ACF).}
To inject 2D semantic priors into the 3D lifting process, we introduce a multi-scale, unidirectional cross-attention fusion module. Crucially, we treat the radiographic features as the "queries" that selectively extract structural guidance from the semantic "keys/values."

\textit{Semantic pyramid alignment.}
Given $I_{\mathrm{PR}}$, the foundation encoder $E_s$ produces a hierarchy of intermediate representations $\{\phi_s^{(l)}\}$. For each scale $l$, we use a lightweight adapter $\mathrm{Proj}_s^{(l)}(\cdot)$ and bilinear interpolation to obtain an aligned semantic feature map:
\begin{equation}
F_{s,\mathrm{map}}^{(l)} \;=\; \mathrm{Interp}\!\left(\mathrm{Proj}_s^{(l)}(\phi_s^{(l)}),\; (H_l,W_l)\right)
\in \mathbb{R}^{C'_l \times H_l \times W_l},
\end{equation}
where $(H_l,W_l)$ follows the progressive decoding pyramid (e.g., $(H,W)$, $(H/2,W/2)$, $(H/4,W/4)$, $(H/8,W/8)$, and $(H/16,W/16)$ in our implementation). Similarly, the radiographic encoder provides radiographic feature maps $F_{r,\mathrm{map}}^{(l)} \in \mathbb{R}^{C_l \times H_l \times W_l}$ .

\textit{Unidirectional cross-attention at scale $l$.}
We reshape the aligned spatial maps into token sequences 
$\mathbf{X}_r^{(l)} \in \mathbb{R}^{N_l \times C_l}$ and 
$\mathbf{X}_s^{(l)} \in \mathbb{R}^{N_l \times C'_l}$, 
where $N_l = H_l W_l$. We then apply standard scaled dot-product cross-attention:
\begin{equation}
X_{\mathrm{attn}}^{(l)} = \mathrm{Attn}\!\left(X_r^{(l)}, X_s^{(l)}, X_s^{(l)}\right),
\end{equation}
where $\mathrm{Attn}(\cdot)$ applies learnable projection with embedding dimension $d_l$.

\textit{Gated residual fusion.}
We regulate the contribution of semantic priors using a learnable scalar gate $\alpha^{(l)}$ initialized to zero:
\begin{equation}
\tilde{X}_r^{(l)} \;=\; X_r^{(l)} + \alpha^{(l)} \cdot X_{\mathrm{attn}}^{(l)}.
\end{equation}
$\tilde{X}_r^{(l)}$ is transposed and reshaped back to a spatial map
$\tilde{F}_{r,\mathrm{map}}^{(l)}$. Finally, we perform channel-wise concatenation with the aligned semantic feature and apply a $1{\times}1$ convolution for channel mixing:
\begin{equation}
F_{\mathrm{fuse}}^{(l)} = \mathrm{Conv}_{1\times1}^{(l)}\!\left([\tilde{F}_{r,\mathrm{map}}^{(l)};\ F_{s,\mathrm{map}}^{(l)}]\right),
\end{equation}

\textit{Progressive decoding.}
The prior-enriched representations $\{F_{\mathrm{fuse}}^{(l)}\}_{l=1}^{L}$ are injected via skip connections into our progressive 3D decoder~\cite{linguraru_3dpx_2024} for the coarse-to-fine synthesis of the canonical volumetric field $\hat{V}_c$.

\subsection{Stage 2: Patient-Specific Anatomical Restoration (AR-Net)}

\begin{algorithm}[t]
\caption{Supervision generation for $P^{*}$}
\label{alg:pstar_generation}
\begin{algorithmic}[1]
\REQUIRE Canonical/native CBCT pair $(V_{c}, V_{n})$
\STATE Estimate dense correspondence between $V_{c}$ and $V_{n}$ under normalized coordinates
\STATE Fit a sparse control lattice $P^{*}\in\mathbb{R}^{3\times D_c\times H_c\times W_c}$ to the dense displacement field
\STATE Regularize the fitted lattice using smoothness and boundary constraints
\STATE Clip extreme control-point displacements to obtain $P^{*}_{\mathrm{clip}}$
\STATE Cache $P^{*}$ and $P^{*}_{\mathrm{clip}}$ for Stage~2 supervision during training
\end{algorithmic}
\end{algorithm}

\label{sec:stage2}
To complete our disentangled PR-to-CBCT inversion paradigm, Stage~2 restores the patient-specific arch geometry intentionally factorized out during Stage~1. Given the canonical volume $\hat{V}_c \in \mathbb{R}^{D\times H\times W}$, AR-Net estimates a differentiable spatial mapping that transfers $\hat{V}_c$ back to the native coordinate system, yielding $\hat{V}_n$. 
We implement this transformation using backward resampling: a dense sampling grid $G \in \mathbb{R}^{D\times H\times W\times 3}$ specifies the corresponding sampling location for each native-space voxel. The restored volume is obtained by:
\begin{equation}
\hat{V}_n(x)=\hat{V}_c\!\left(G(x)\right),
\end{equation}.

\textbf{Coarse arch restoration via prior-initialized control-point grids.}
Direct regression of a full-resolution 3D sampling grid is highly ill-conditioned. To constrain the search space, we parameterize $G$ using a sparse control-point lattice
$P \in \mathbb{R}^{3 \times D_c \times H_c \times W_c}$. The lattice is initialized with an analytical arch-shaped prior $P_0$ (e.g., parabolic/cylindrical), which provides a weak geometric bias toward plausible mandibular configurations while remaining patient-adaptive~\cite{sederberg_free-form_1986,kanitsar_cpr_2002,rueckert_nonrigid_1999,alharbi_mathematical_2008,muhamad2015curve}. By serving as a sensible warm start, this initialization significantly constrains the severely ill-posed inverse mapping. CPNet3D, a lightweight convolutional network, predicts residual updates on this lattice conditioned on a downsampled $\hat{V}_c$:
\begin{equation}
P_{\mathrm{coarse,raw}} = P_0 + \Delta P_{\mathrm{coarse}}.
\end{equation}

A dense coarse grid is obtained via trilinear interpolation, $G_{\mathrm{coarse}} = \mathrm{Up}\!\left(\mathrm{clip}(P_{\mathrm{coarse,raw}})\right),$ yeilding the globally aligned intermediate reconstruction $\hat{V}_{n}^{\mathrm{coarse}} = \mathcal{W}\!\left(\hat{V}_c,\ G_{\mathrm{coarse}}\right).$

\textbf{Fine refinement by residual control-lattice update.}
To captures localized morphological nuances, RefineNet3D predicts a fine residual update using a a joint representation of the canonical reference and the coarse estimate:
\begin{equation}
\Delta P_{\mathrm{fine}} = \mathrm{RefineNet3D}\!\left([\hat{V}_c;\ \hat{V}_{n}^{\mathrm{coarse}};\ \tilde{G}_{\mathrm{coarse}}]\right),
\end{equation}
The refined control lattice $P_{\mathrm{final,raw}} = P_{\mathrm{coarse,raw}} + \Delta P_{\mathrm{fine}}$ is converted to the final dense grid $G_{\mathrm{final}}$, yeilding the ultimate native-space reconstruction $\hat{V}_n = \mathcal{W}\!\left(\hat{V}_c,\ G_{\mathrm{final}}\right).$

\textbf{Training objective.}
We supervise the predicted control lattice in absolute normalized coordinates using the ground-truth lattice $P^{\ast}$. The training-only generation pipeline of $P^{\ast}$ is summarized in Algorithm~\ref{alg:pstar_generation}. Hard clamping is applied \emph{only} during warping to satisfy the valid sampling domain of \texttt{grid\_sample}, while the learning signal is imposed on the raw prediction:
\begin{equation}
\mathcal{L}_{\mathrm{raw}} = \left\lVert P_{\mathrm{final,raw}} - P^{\ast} \right\rVert_1 .
\end{equation}
We also apply a clipped-consistency term on the warped-domain coordinates, 
$\mathcal{L}_{\mathrm{clip}} = \left\lVert \mathrm{clip}(P_{\mathrm{final,raw}}) \right\rVert$, 
alongside smoothness (TV) and out-of-bound penalties to mitigate unrealistic folding.
\begin{equation}
\begin{aligned}
\mathcal{L} &= \mathcal{L}_{\mathrm{raw}} + \lambda_{\mathrm{clip}}\,\mathcal{L}_{\mathrm{clip}} + \lambda_{\mathrm{tv}}\,\mathrm{TV}(P_{\mathrm{final,raw}}) \\
&\quad + \lambda_{\mathrm{oob}}\,\left\lVert 
\max(|P_{\mathrm{final,raw}}|-1,0) 
\right\rVert_1
\end{aligned}
\end{equation}

\section{Experimental Setup}
\subsection{Datasets}

We evaluate the proposed framework on three publicly available CBCT datasets with diverse acquisition protocols and spatial resolutions: the Cui22 dataset (NC)~\cite{cui_fully_2022}, the MMDental dataset (ST)~\cite{wang_mmdental-multimodal_2025}, and the ToothFairy dataset (TF)~\cite{lumetti_enhancing_2024,bolelli_segmenting_2024,cipriano_improving_2022}. Each dataset was treated as an independent benchmark and split patient-wise into training, validation, and test sets using an 80:10:10 ratio. The selected datasets provide a comprehensive testbed for clinical evaluation with their main characteristics summarized in Table~\ref{tab:dataset_stats}. For the NC dataset, we followed prior work to exclude scans with severe metal artifacts or incomplete fields of view from its original 100 subjects. The ST dataset introduces challenging real-world scenarios with widespread metallic restorations, while the TF dataset provides high-quality scans naturally aligned with our target dentoalveolar field of view.

\begin{table}[ht]
\centering
\caption{Dataset statistics after preprocessing. NC and ST scans were cropped to match the TF dentoalveolar field of view (FOV). All volumes were resampled to $128\times128\times256$.}
\label{tab:dataset_stats}
\begin{tabular}{lccc}
\toprule
Dataset & Native Res. (mm) & Input FOV & Subjects \\
\midrule
NC~\cite{cui_fully_2022} & 0.40 & Cropped & 89 \\
ST~\cite{wang_mmdental-multimodal_2025} & 0.25 & Cropped & 202 \\
TF~\cite{lumetti_enhancing_2024,bolelli_segmenting_2024,cipriano_improving_2022} & 0.30 & Native & 63 \\
\bottomrule
\end{tabular}
\end{table}

\begin{table*}[ht]
\caption{Comparison with state-of-the-art 2D-to-3D reconstruction methods on the NC, ST, and TF datasets.}
\label{tab:sota_comparison}
\centering
\setlength{\tabcolsep}{3pt}
\footnotesize

\begin{tabular}{lccccccccc}
\hline
\multirow{2}{*}{Model} & \multicolumn{3}{c}{NC} & \multicolumn{3}{c}{ST} & \multicolumn{3}{c}{TF} \\
\cline{2-10}
& PSNR(dB) & DSC(\%) & SSIM(\%) & PSNR(dB) & DSC(\%) & SSIM(\%) & PSNR(dB) & DSC(\%) & SSIM(\%) \\
\hline
UNETR~\cite{hatamizadeh_unetr_2022}
& 15.07 & 60.14$^{*}$ & 62.14
& 16.42 & 61.84$^{*}$ & 69.28
& 17.25 & 61.85$^{*}$ & 62.25 \\
Oral-3D~\cite{song_oral-3d_2021}
& 20.41 & 63.13$^{*}$ & 69.78
& 22.27 & 68.52$^{*}$ & 69.49
& 17.22 & 68.76$^{*}$ & 64.33 \\
3DPX~\cite{linguraru_3dpx_2024}
& \underline{20.46} & \underline{70.33$^{*}$} & \underline{73.08}
& \underline{23.23} & \underline{73.91$^{*}$} & \underline{72.34}
& \underline{20.25} & \underline{77.87$^{*}$} & \underline{75.29} \\
\textbf{HyDAR-Pano3D}
& \textbf{25.76} & \textbf{76.47} & \textbf{80.33}
& \textbf{24.53} & \textbf{74.63} & \textbf{74.09}
& \textbf{22.98} & \textbf{83.83} & \textbf{85.70} \\
\hline
\end{tabular}

\parbox{0.98\textwidth}{\footnotesize
PSNR, DSC, and SSIM are reported in dB or percent as appropriate; higher values indicate better performance. Bold and underlined values denote the best and second-best results, respectively.
$^{*}$ indicates $p<0.05$ versus HyDAR-Pano3D based on a paired $t$-test on per-case DSC.}
\end{table*}

To standardize the anatomical coverage and construct paired training data, we applied a unified preprocessing pipeline. For the NC and ST datasets, full-craniofacial CBCT scans were first cropped to the dentoalveolar region using bounding boxes derived from the available segmentation masks. This step standardized the anatomical field of view to better match the TF dataset by retaining the complete maxillomandibular dentoalveolar structures while excluding unrelated craniofacial regions. Subsequently, all cropped volumes from the three datasets were resampled to a fixed size of $128 \times 128 \times 256$ ($D \times H \times W$) using trilinear interpolation. Note that mask-based ROI extraction was utilized only during data set preprocessing and is not required during model training or inference.

Because paired clinical PR and CBCT volumes are rarely available, we generated synthetic PRs directly from CBCT volumes to serve as anatomically aligned paired training data. Specifically, curve-guided curved planar reformation (CPR)~\cite{kanitsar_cpr_2002} was performed using manually annotated dental arch curves, followed by mean intensity projection along the arch-aligned depth axis. The necessary dental arch curves were annotated by a trained medical imaging researcher under the supervision of a Lecturer and Specialist in Dento-Maxillofacial Radiology. We explicitly note that these synthetic panoramas are derived via CPR and inherently differ from conventional clinical PRs based on linear tomography principles~\cite{kwon_panoramic_2023}. Specifically, they do not capture the highly complex anatomical overlap and non-uniform magnification present in real PR acquisitions. Nevertheless, they produce anatomically consistent panoramic projections that strictly preserve spatial correspondence with the source CBCT volumes. Finally, these manual annotations were used exclusively during offline synthetic data preparation and are completely bypassed during actual model training and inference.

\section{Results and Discussion}
\label{sec:results_discussion}

\subsection{Experimental Protocol and Implementation Details}
All experiments were conducted independently on each dataset without cross-dataset mixing. Within each dataset, all internal variants and external baselines were trained and evaluated under identical data splits and preprocessing settings to ensure fair comparisons. To investigate the effect of semantic priors in canonical reconstruction, we evaluated three foundation-model encoders within the dual-encoder framework: SAM-ViT-B~\cite{kirillov_segment_2023}, MedSAM~\cite{ma_segment_2024}, and SemiT-SAM~\cite{wang_semit-sam_2025}. During training, pretrained backbone parameters were fine-tuned using a reduced learning rate ($0.1\times$ that of the remaining network parameters). All models were implemented in PyTorch and trained on NVIDIA A6000 GPUs using the AdamW optimizer with a batch size of 4, an initial learning rate of $1\times10^{-4}$, and weight decay of $1\times10^{-4}$; cosine annealing with linear warm-up was applied during optimization. Stage~1 was trained for 200 epochs to learn canonical volumetric reconstruction, and Stage~2 was subsequently trained for 200 epochs in a sequential manner by first optimizing the coarse deformation module and then the refinement module while freezing previously trained components. Random 3D rotations ($\pm3^\circ$, $\pm5^\circ$, and $\pm10^\circ$) were applied as data augmentation across all datasets. To isolate the contribution of Stage~1 components, we evaluated four architectural variants: FM-Only (retaining only the foundation model branch), 3DPX-Only (retaining only the geometry-aware 3DPX branch), Concatenation (replacing adaptive cross-attention fusion with simple feature concatenation), and Full HyDAR-Pano3D (the complete dual-encoder architecture with ACF-based fusion). Unless otherwise specified, all variants shared the same Stage~2 deformation module.

\begin{figure}[ht]
    \centering
    \includegraphics[width=0.9\linewidth]{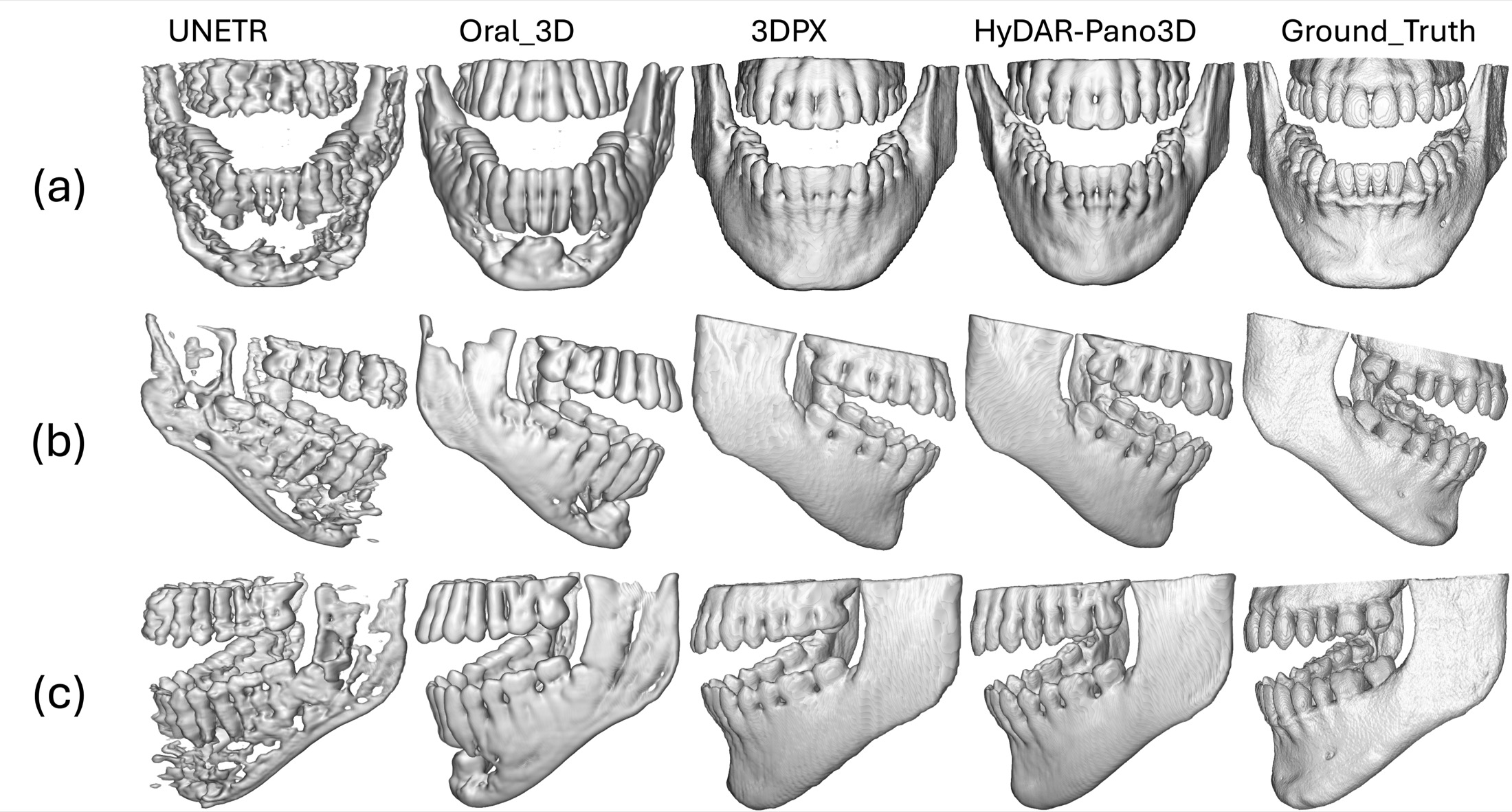}
    \caption{Qualitative comparison of panoramic 2D-to-3D dental reconstruction on representative TF cases. Reconstructions from \textsc{UNETR}~\cite{hatamizadeh_unetr_2022}, \textsc{Oral-3D}~\cite{song_oral-3d_2021}, \textsc{3DPX}~\cite{linguraru_3dpx_2024}, and \textsc{HyDAR-Pano3D} are contrasted with the ground-truth CBCT volumes. Rows (a)--(c) present different cases under multiple viewpoints.}
    \label{fig:toothfairy_recons}
\end{figure}

\subsection{Evaluation Metrics}
\label{sec:metrics}

Reconstruction performance was evaluated using Peak Signal-to-Noise Ratio (PSNR), Structural Similarity Index (SSIM), and Dice Similarity Coefficient (DSC), following prior work on panoramic 2D-to-3D reconstruction~\cite{song_oral-3d_2021}. PSNR and SSIM quantify voxel-level fidelity and structural similarity between reconstructed and ground-truth volumes. For DSC, the reconstructed volume $V_{\text{pred}}$ and ground-truth volume $V_{\text{gt}}$ were first independently min--max normalized to $[0,1]$, and a shared threshold $\mu_{gt}$ (the mean intensity of normalized $V_{\text{gt}}$) was used to construct binary masks:
\[
M_{\text{pred}} = \mathbf{1}(V_{\text{pred}} > \mu_{gt}), \qquad
M_{\text{gt}}   = \mathbf{1}(V_{\text{gt}} > \mu_{gt}),
\]

\begin{table*}[t]
\caption{Ablation study on encoder configurations and fusion strategies across the NC, ST, and TF datasets.}
\label{tab:ablation_summary}
\centering
\setlength{\tabcolsep}{3pt}
\renewcommand{\arraystretch}{1.05}
\footnotesize

\begin{tabular}{lccccccccccccc}
\hline
\multicolumn{14}{c}{\textbf{(a) Encoder configurations}} \\
\hline
\multirow{2}{*}{Model}
& \multicolumn{4}{c}{Encoders}
& \multicolumn{3}{c}{NC}
& \multicolumn{3}{c}{ST}
& \multicolumn{3}{c}{TF} \\
\cline{2-14}
& \textit{S.} & \textit{M.} & \textit{Se.} & \textit{C.}
& PSNR(dB) & DSC(\%) & SSIM(\%)
& PSNR(dB) & DSC(\%) & SSIM(\%)
& PSNR(dB) & DSC(\%) & SSIM(\%) \\
\hline

\multirow{3}{*}{Single-Encoder}
& \checkmark &         &         &
& 21.06 & 72.30 & 73.20
& 22.19 & 71.03 & 71.26
& 21.02 & 71.39 & 73.56 \\
&         & \checkmark &         &
& 21.62 & 72.70 & 74.03
& 22.38 & 71.16 & 71.84
& \underline{21.58} & 73.53 & 72.72 \\
&         &         & \checkmark &
& \textbf{25.64} & 73.90 & 75.91
& 23.87 & 72.30 & 72.73
& \textbf{21.59} & \underline{82.33} & \underline{82.58} \\
\hline

\multirow{3}{*}{Dual-Encoder (Concat)}
& \checkmark &         &         & \checkmark
& 24.82 & \textbf{75.38} & \textbf{78.72}
& \underline{24.25} & 72.42 & \textbf{73.74}
& 21.05 & 74.97 & 75.52 \\
&         & \checkmark &         & \checkmark
& \underline{24.99} & 71.37 & \underline{76.11}
& \textbf{24.99} & \textbf{73.08} & \underline{73.71}
& 21.55 & 75.92 & 75.52 \\
&         &         & \checkmark & \checkmark
& 24.42 & \underline{75.32} & 75.76
& 24.01 & \underline{72.83} & 72.75
& 21.06 & \textbf{83.63} & \textbf{84.12} \\
\hline
\end{tabular}


\begin{tabular}{ccccccccccccc}
\hline
\multicolumn{13}{c}{\textbf{(b) Fusion strategy: concatenation vs.\ ACF}} \\
\hline
\multicolumn{3}{c}{Encoder} & Fusion
& \multicolumn{3}{c}{NC}
& \multicolumn{3}{c}{ST}
& \multicolumn{3}{c}{TF} \\
\hline
\textit{S.} & \textit{M.} & \textit{Se.} &
& PSNR(dB) & DSC(\%) & SSIM(\%)
& PSNR(dB) & DSC(\%) & SSIM(\%)
& PSNR(dB) & DSC(\%) & SSIM(\%) \\
\hline

\checkmark &         &         & Concat
& 24.82 & 75.38 & \underline{78.72}
& 24.25 & 72.42 & 73.74
& 21.05 & 74.97 & 75.52 \\

\checkmark &         &         & ACF
& 25.52 & 75.38 & 78.69
& \textbf{25.07} & 73.01 & \textbf{74.27}
& 21.68 & 82.55 & 80.38 \\
\hline

         & \checkmark &         & Concat
& 24.99 & 71.37 & 76.11
& \underline{24.99} & 73.08 & 73.74
& 21.55 & 75.92 & 75.52 \\
         & \checkmark &         & ACF
& \underline{25.53} & \underline{75.90} & 75.91
& 24.38 & \underline{73.21} & 73.84
& \underline{21.89} & 80.25 & 83.08 \\
\hline

         &         & \checkmark & Concat
& 24.42 & 75.32 & 75.76
& 24.01 & 72.83 & 72.75
& 21.06 & \underline{83.63} & \underline{84.12} \\
         &         & \checkmark & ACF
& \textbf{25.76} & \textbf{76.47} & \textbf{80.33}
& 24.53 & \textbf{74.63} & \underline{74.09}
& \textbf{22.98} & \textbf{83.83} & \textbf{85.70} \\
\hline
\end{tabular}

\parbox{0.98\textwidth}{\footnotesize
S.\ = SAM-ViT-B, M.\ = MedSAM-lite, Se.\ = SemiT-SAM, C.\ = concatenation, and ACF = adaptive cross-attention fusion.}

\parbox{0.98\textwidth}{\footnotesize
PSNR, DSC, and SSIM are reported in dB or percent as appropriate; higher values indicate better performance. Bold and underlined values denote the best and second-best results, respectively. $^{*}$ indicates $p<0.05$ versus HyDAR-Pano3D based on a paired $t$-test on per-case DSC.}
\end{table*}

where $\mathbf{1}(\cdot)$ is the indicator function. DSC was computed between $M_{\text{pred}}$ and $M_{\text{gt}}$; thus, this evaluation mainly reflects the overlap of high-density dentoalveolar structures (e.g., teeth and jawbone) while penalizing relative intensity miscalibration.

Statistical significance was assessed using a paired t-test, where $^{*}$ indicates $p<0.05$.

\subsection{Comparison with State-of-the-Art Methods}

Due to the inherent ill-posed nature of single-view dental reconstruction and the limited availability of fully open-source architectures in this domain, we established a comprehensive evaluation benchmark using three representative baselines. These include: (1) UNETR \cite{hatamizadeh_unetr_2022}, serving as the established baseline for general-purpose 3D medical vision architectures; (2) Oral-3D \cite{song_oral-3d_2021}, a widely recognized domain-specific framework that introduces dedicated spatial priors for dental modeling; and (3) \textsc{3DPX} \cite{linguraru_3dpx_2024}, the most recent state-of-the-art method, included to contextualize the performance gains of our proposed framework. We evaluated these models on three public datasets: NC, ST, and TF. In all experiments, our HyDAR-Pano3D employed the SemiT-SAM FM encoder, as it demonstrated the best overall performance among the evaluated semantic-prior variants.

\paragraph{Overall performance.}
Quantitative Comparison. Across all datasets, HyDAR-Pano3D consistently outperforms existing methods (Table~\ref{tab:sota_comparison}). On the high-resolution TF dataset, our approach achieves superior DSC (83.83\%) and SSIM (85.70\%), reflecting the precise preservation of tooth morphology and volumetric continuity. Similar performance gains are observed on NC (25.76dB PSNR, 80.33\% DSC) and ST (24.53dB PSNR, 74.09\% DSC); despite the latter's prevalent metal artifacts, HyDAR-Pano3D maintains a significant lead over the strongest baseline (3DPX), demonstrating robustness across diverse imaging conditions and multi-vendor scans.

Qualitative Evaluation. Visual comparisons on the TF dataset (Fig.~\ref{fig:toothfairy_recons}) demonstrate that HyDAR-Pano3D achieves a marked reduction in structural blurring and produced highly consistent tooth-root reconstructions compared to baseline methods—attributes that are essential for reliable downstream clinical analysis. Although competing approaches frequently exhibit surface irregularities and contour distortions, our framework recovers a smoother, anatomically faithful geometry. These qualitative improvements confirm that decoupling canonical space inference from spatial restoration effectively mitigates the inherent ambiguities and artifact interference typical of single-view reconstruction.

\paragraph{Statistical considerations.}

We report the mean performance among the test subjects. Paired significance testing required per-subject outputs from all compared baselines under identical protocols; however, such outputs are not available for all methods. The results reported for the state-of-the-art methods in Table III were obtained after applying the AR-Net, which deforms the flattened CBCT predictions into the native CBCT space of the ground-truth data. To further examine whether the observed improvements arise from the generative stage rather than from the subsequent geometric transformation, we also evaluated the intermediate flattened CBCT predictions without AR-Net. This evaluation produced consistent findings with the deformed results, with our method achieving an 11.2\% higher DSC than the second-best method, 3DPX. To ensure a fair comparison under consistent geometric assumptions, we therefore focus our ablation studies on components within the proposed two-stage pipeline. A direct “no-canonical” variant is not considered, as panoramic radiographs are inherently acquired in an arch-aligned projection domain; removing the canonical stage would alter the problem formulation rather than isolate the contribution of an individual architectural component.

\subsection{Ablation Analysis}
To analyze the contributions of individual components in the proposed framework, we conduct ablation studies on NC, ST, and ToothFairy. Results are summarized in Table~\ref{tab:ablation_summary}.

\paragraph{Single- vs. dual-encoder designs.}
In the single-encoder setting, replacing a purely convolutional encoder with foundation-model backbones consistently improved reconstruction quality across datasets, indicating that large-scale pretrained semantic priors provide more stable anatomical cues from panoramic radiographs.
Among the evaluated backbones, SemiT-SAM~\cite{wang_semit-sam_2025} yielded the strongest and most consistent performance, suggesting that domain-adapted semantic representations are particularly beneficial for dental anatomy.

In the dual-encoder setting, combining a convolutional encoder with a foundation-model encoder further improved reconstruction performance.
This configuration consistently outperformed either encoders alone, indicating that local geometric cues captured by convolutional features and global semantic context provided by foundation models are complementary for canonical anatomical inference.

\paragraph{Fusion strategy: ACF vs. concatenation.}
We further compared the proposed ACF with simple feature concatenation. Across encoder configurations and datasets, ACF consistently yields higher PSNR, SSIM, and DSC. This suggests that explicitly modeling interactions between radiographic and semantic features facilitates more effective integration than naive concatenation, particularly for preserving fine-grained anatomical structures. The improvements are most pronounced on ToothFairy, where detailed tooth morphology places higher demands on accurate feature alignment.

\subsection{Downstream segmentation}

We evaluated 3D segmentation performance on reconstructed CBCT volumes to assess whether the proposed reconstruction framework preserves sufficient anatomical fidelity for clinically relevant analysis. Specifically, we adopted segmentation of teeth on both the TF and NC datasets, and segmentation of the IAC on the TF dataset, using the same pretrained 3D segmentation model across all reconstruction methods, without any fine-tuning.

\begin{table}[t]
\caption{Comparison of downstream IAC segmentation performance on the TF dataset using reconstructed CBCT volumes.}
\label{tab:iac}
\centering
\setlength{\tabcolsep}{6pt}
\renewcommand{\arraystretch}{1.05}
\footnotesize

\begin{tabular}{lcc}
\hline
Method & DSC(\%)$\uparrow$ & HD95(mm)$\downarrow$ \\
\hline
UNETR~\cite{hatamizadeh_unetr_2022} & 49.5 & 12.82 \\
Oral-3D~\cite{song_oral-3d_2021}    & 53.4 & 11.44 \\
3DPX~\cite{linguraru_3dpx_2024}     & \underline{62.3} & \underline{5.37} \\
HyDAR-Pano3D                 & \textbf{72.2} & \textbf{2.96} \\
\hdashline
\textit{Oracle (Real CBCT)} & \textit{95.9} & \textit{0.32} \\

\hline
\end{tabular}

\parbox{0.98\columnwidth}{\footnotesize
DSC and HD95 denote Dice similarity coefficient and 95th percentile Hausdorff distance, respectively. Higher DSC and lower HD95 indicate better performance. Bold values denote the best results.}
\end{table}

\begin{figure}[ht]
    \centering
    \includegraphics[width=\linewidth]{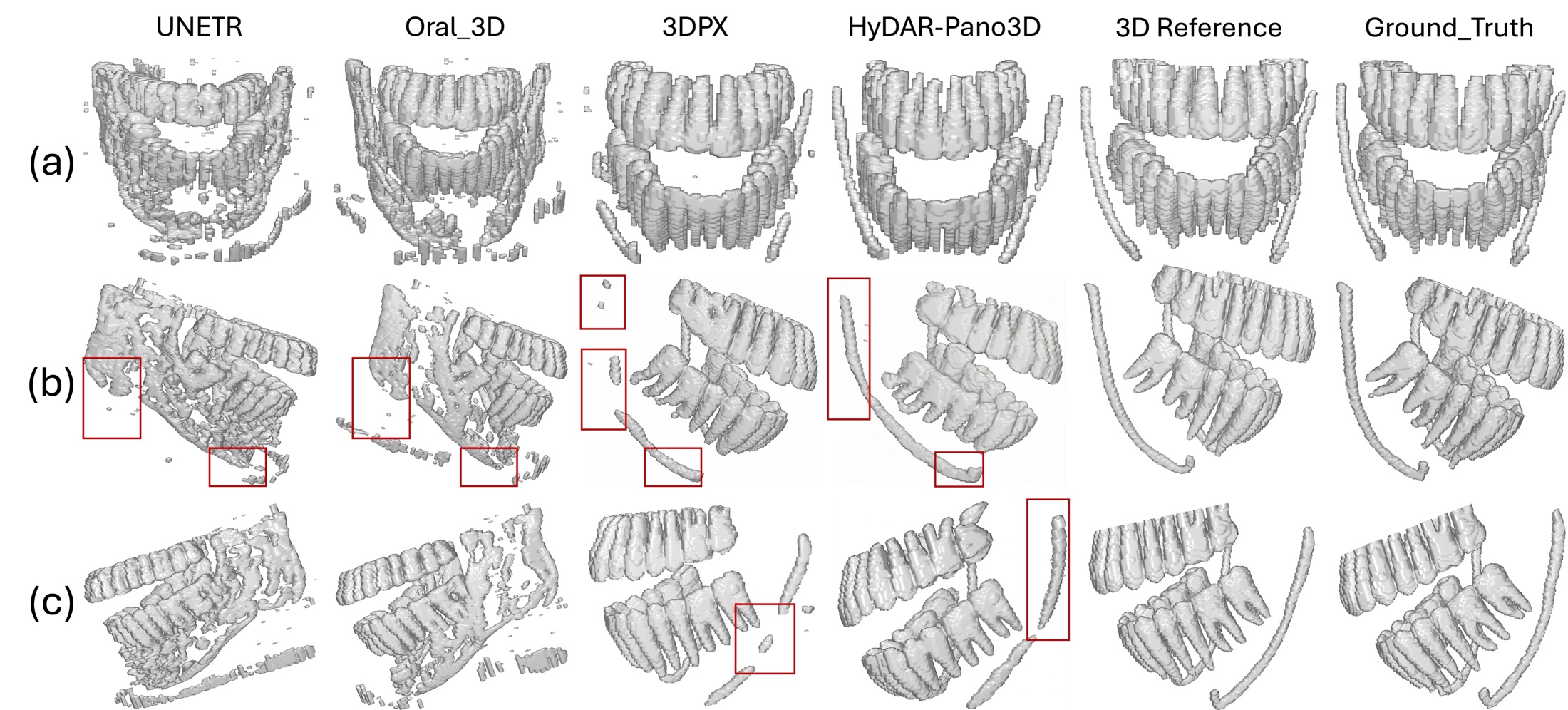}
    \caption{Qualitative comparison of downstream tooth and inferior alveolar canal (IAC) segmentation on the TF dataset. Reconstructed CBCT volumes from various methods (UNETR, Oral-3D, 3DPX, and our HyDAR-Pano3D) were fed into the same pretrained 3D segmentation network without fine-tuning. The fully supervised 3D reference represents the target performance obtained by applying the identical nnU-Net segmentation model directly to the real clinical CBCTs. Ground Truth denotes the manual expert annotations. (a)–(c) show representative test cases. Red boxes highlight regions with severe structural discrepancies or topological breaks (e.g., disconnected IACs or missing molars).}
    \label{fig:iac}
\end{figure}

Quantitatively, reconstructions produced by HyDAR-Pano3D consistently yielded higher Dice scores and lower HD95 errors compared to competing 2D-to-3D methods across all evaluated settings (Tables~\ref{tab:iac} and~\ref{tab:seg}). Most notably, on the TF dataset, HyDAR-Pano3D achieved a substantial DSC of 72.2\% for Inferior Alveolar Canal (IAC) segmentation, outperforming the strongest baseline (\textsc{3DPX}) by a significant margin of approximately 10\% (Table~\ref{tab:iac}). Because the IAC is a thin, elongated tubular structure embedded within the mandible, it is highly sensitive to geometric discontinuities and local deformation errors. This specific leap in accuracy explicitly confirms that our framework successfully preserves geometric continuity and boundary integrity.

\begin{table}[t]
\caption{Downstream tooth segmentation performance on the TF and NC datasets using reconstructed CBCT volumes from different 2D-to-3D methods.}
\label{tab:seg}
\centering
\setlength{\tabcolsep}{4pt}
\footnotesize
\begin{tabular}{lcccc}
\hline
 & \multicolumn{2}{c}{TF} & \multicolumn{2}{c}{NC} \\
\hline
Method & DSC (\%)$\uparrow$ & HD95 (mm)$\downarrow$ & DSC (\%)$\uparrow$ & HD95 (mm)$\downarrow$ \\
\hline
UNETR~\cite{hatamizadeh_unetr_2022}  & 54.4 & 4.51 & 49.1 & 11.01 \\
Oral-3D~\cite{song_oral-3d_2021}     & 71.9 & 3.53 & 54.2 & 8.96 \\
3DPX~\cite{linguraru_3dpx_2024}      & \underline{75.1} & \underline{3.09} & \underline{64.6} & \underline{5.42} \\
HyDAR-Pano3D               & \textbf{82.4} & \textbf{2.68} & \textbf{76.4} & \textbf{2.56} \\
\hdashline
\textit{Oracle} & \textit{92.6} & \textit{4.49} & \textit{89.7} & \textit{1.67}\\
\hline
\end{tabular}

\parbox{0.98\columnwidth}{\footnotesize
DSC and HD95 denote Dice similarity coefficient and 95th percentile Hausdorff distance, respectively. Higher DSC and lower HD95 indicate better performance. Bold values denote the best results.}
\end{table}

Furthermore, consistent quantitative gains were observed for tooth segmentation in the NC dataset (Table~\ref{tab:seg}). Although macroscopic tooth segmentation is less susceptible to fine geometric errors than the IAC, these results demonstrate that the structural benefits of our framework generalize across anatomical targets of varying morphological complexity.

\begin{figure}
    \centering
    \includegraphics[width=\linewidth]{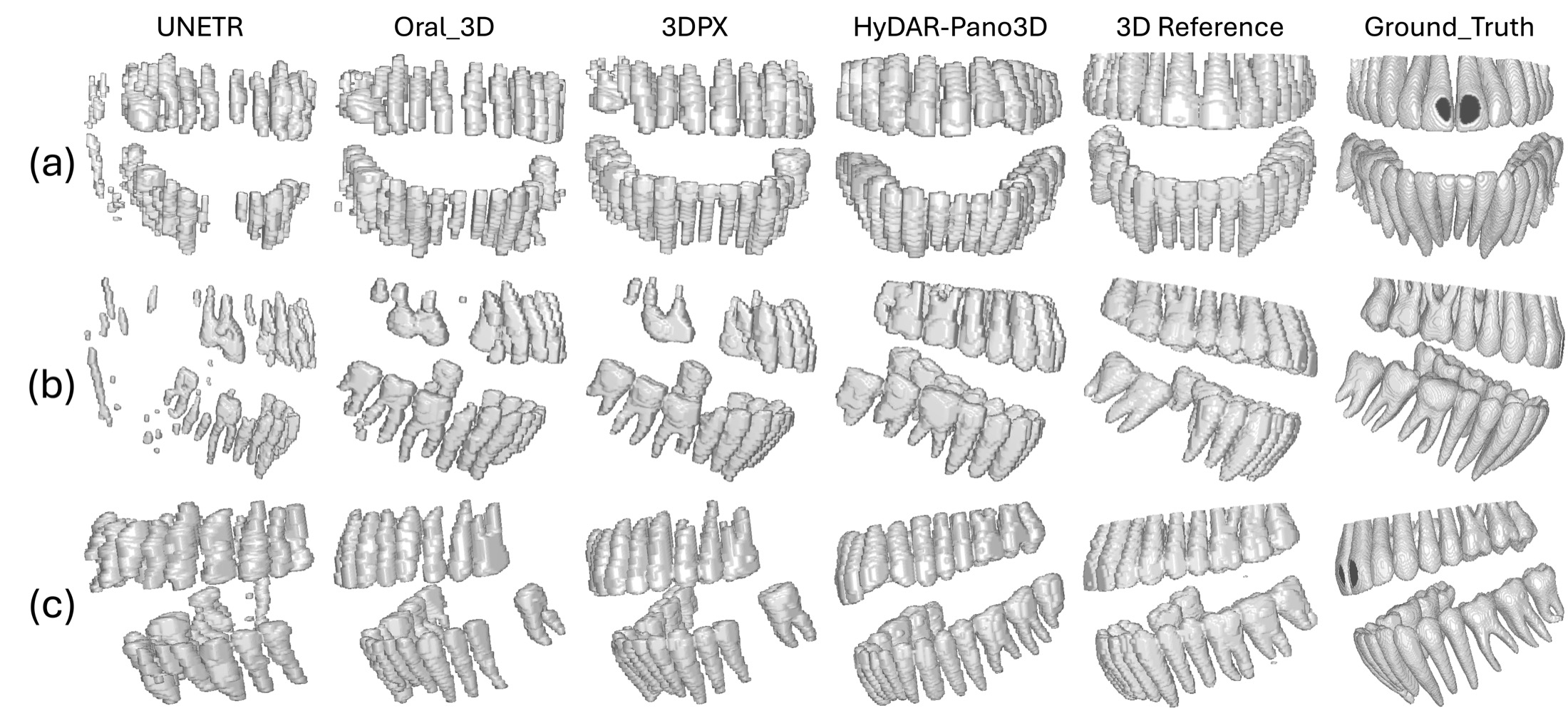}
    \caption{Qualitative comparison of downstream tooth segmentation on the TF dataset. Reconstructed CBCT volumes from various methods (UNETR, Oral-3D, 3DPX, and our HyDAR-Pano3D) were segmented using the same pretrained network without fine-tuning. The fully supervised 3D reference represents the target performance obtained by applying the identical nnU-Net segmentation model directly to the real clinical CBCTs. Ground Truth denotes the manual expert annotations. (a)–(c) show representative cases.}
    \label{fig:tf_recons}
\end{figure}

\begin{figure}
    \centering
    \includegraphics[width=\linewidth]{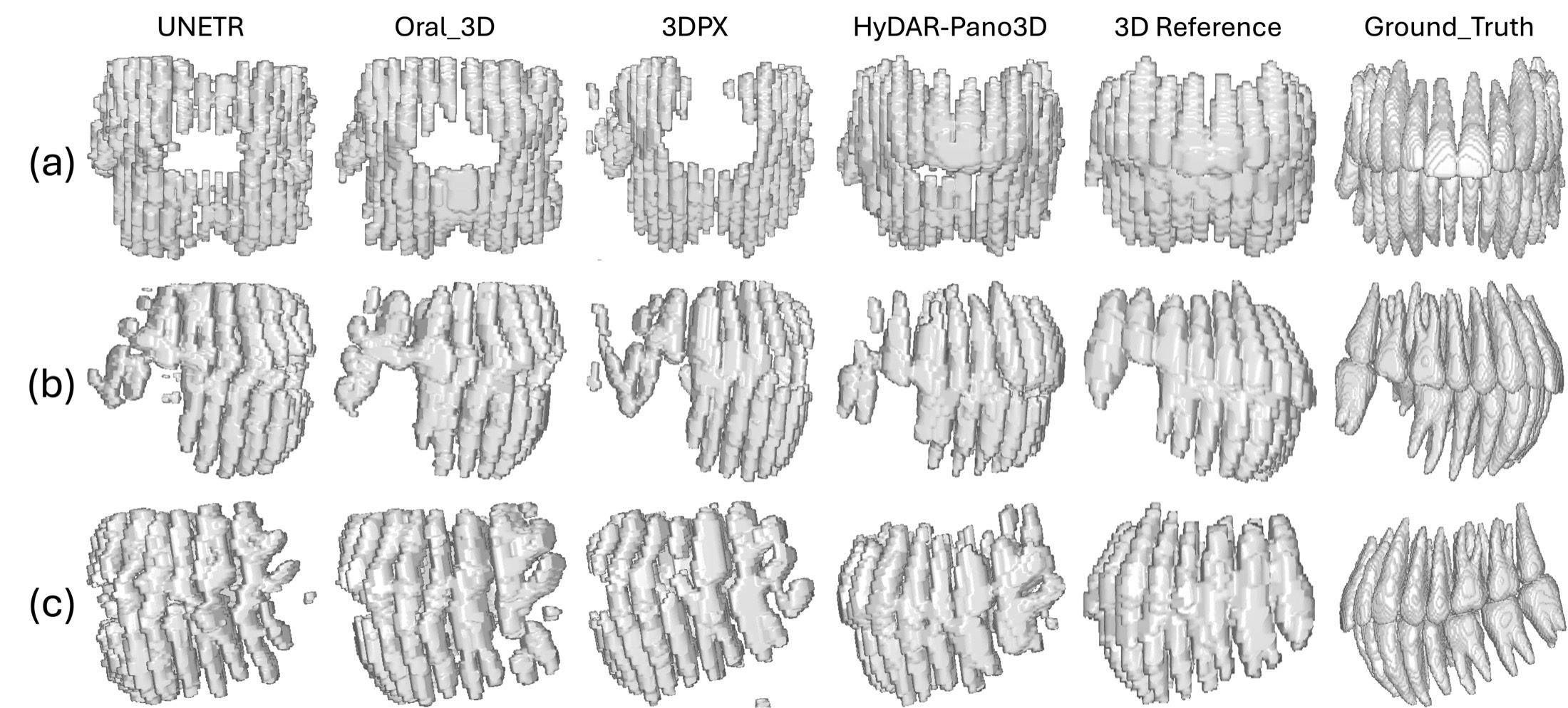}
    \caption{Qualitative comparison of downstream tooth segmentation on the NC dataset. Reconstructed CBCT volumes from various methods (UNETR, Oral-3D, 3DPX, and our HyDAR-Pano3D) were segmented using the same pretrained network without fine-tuning. The fully supervised 3D reference represents the target performance obtained by applying the identical nnU-Net segmentation model directly to the real clinical CBCTs. Ground Truth denotes the manual expert annotations. (a)–(c) show representative cases.}
    \label{fig:nc_recons}
\end{figure}

Together, these results indicate that improved downstream performance is closely associated with higher reconstruction performance rather than differences in segmentation modeling. By disentangling canonical anatomical density inference from anatomical restoration via deformation learning, HyDAR-Pano3D produce reconstructions that form a more reliable volumetric substrate for automated dental analysis.

\subsection{Limitations and Future Work}
Despite the encouraging results, several limitations of the proposed framework should be acknowledged.

First, the supervised training of HyDAR-Pano3D relied on synthetically generated panoramic radiographs derived from CBCT volumes. Although the adopted curve-guided projection strategy captured key characteristics of panoramic imaging, including projection overlap and non-linear distortion, it may not have fully reflected all scanner-specific acquisition effects present in real clinical panoramic radiographs.
Bridging the remaining domain gap between synthetic and real PR images remains an open challenge.

Second, our framework requires a reasonably accurate arch curve to generate the flattened 3D volumes. We clarify that this manual annotation is performed exclusively during the offline data-preprocessing stage to prepare the ground-truth data. It is completely bypassed during the network training and inference, though it does introduce additional data preparation effort.

Finally, the proposed deformation-based restoration focused on modeling smooth, anatomically plausible geometric variation. Although not observed across all three datasets that represent clinical variabilities, extreme anatomical abnormalities or severe pathological deformations that deviate substantially from the canonical prior may not be fully recovered.
In such cases, the predicted deformation field may oversmooth highly localized irregularities.

Several promising directions can extend our current work. Foremost, explicitly advancing the simulation of projection overlaps and non-linear distortions serves as a critical future step to bridge the aforementioned gap between clinical and synthetic PRs. Additionally,  to further reduce reliance on synthetic supervision via domain adaptation or self-supervised learning for better generalization to real-world panoramic radiographs from diverse scanners and protocols, replacing manual dental arch annotation with automated or learnable arch estimation to enable fully data-driven and scalable canonical-space construction, moving beyond static single-image reconstruction to uncertainty-aware or probabilistic modeling that better captures single-view ambiguity, and integrating the framework with downstream clinical tasks (e.g., implant planning and nerve risk assessment) in end-to-end learning or decision-support pipelines for real-world deployment.

\section*{References}
\bibliographystyle{IEEEtran}
\bibliography{references}

@article{kwon_panoramic_2023,
  title     = {Panoramic dental tomosynthesis imaging by use of {CBCT} projection data},
  author    = {Kwon, Taejin and Choi, Da-in and Hwang, Jaehong and Lee, Taewon and Lee, Inje and Cho, Seungryong},
  journal   = {Sci. Rep.},
  volume    = {13},
  number    = {1},
  pages     = {8817},
  year      = {2023},
  publisher = {Nature Publishing Group UK London}
}

@article{sederberg_free-form_1986,
  title   = {Free-form deformation of solid geometric models},
  volume  = {20},
  number  = {10},
  author  = {Sederberg, Thomas W. and Parry, Scott R.},
  pages   = {151--160},
  journal = {SIGGRAPH Comput. Graph.},
  year    = {1986}
}

@inproceedings{kanitsar_cpr_2002,
  title     = {{CPR} - curved planar reformation},
  booktitle = {Proc. {IEEE} Visualization ({VIS})},
  publisher = {IEEE},
  author    = {Kanitsar, A. and Fleischmann, D. and Wegenkittl, R. and Felkel, P. and Groller, E.},
  year      = {2002},
  pages     = {37--44}
}

@article{muhamad2015curve,
  title   = {The curve of dental arch in normal occlusion},
  author  = {Muhamad, Abu-Hussein and Nezar, Watted and Azzaldeen, Abdulgani},
  journal = {Open Sci. J. Clin. Med.},
  volume  = {3},
  number  = {2},
  pages   = {47--54},
  year    = {2015}
}

@article{alharbi_mathematical_2008,
  title   = {Mathematical analyses of dental arch curvature in normal occlusion},
  volume  = {78},
  number  = {2},
  journal = {Angle Orthod.},
  author  = {AlHarbi, Seba and Alkofide, Eman A. and AlMadi, Abdulaziz},
  year    = {2008},
  pages   = {281--287}
}

@article{rueckert_nonrigid_1999,
  title   = {Nonrigid registration using free-form deformations: application to breast {MR} images},
  volume  = {18},
  number  = {8},
  journal = {IEEE Trans. Med. Imaging},
  author  = {Rueckert, D. and Sonoda, L.I. and Hayes, C. and Hill, D.L.G. and Leach, M.O. and Hawkes, D.J.},
  year    = {1999},
  pages   = {712--721}
}

@article{feng_utility_nodate,
  title   = {On the {Utility} of {Foundation} {Models} for {Fast} {MRI}: {Vision}-{Language}-{Guided} {Image} {Reconstruction}},
  author  = {Feng, Ruimin and He, Xingxin and Mercer, Ronald and Stewart, Zachary and Liu, Fang},
  journal = {arXiv preprint arXiv:2511.19641},
  year    = {2025}
}

@inproceedings{morshuis_segmentation-guided_2024,
  title     = {Segmentation-guided {MRI} reconstruction for meaningfully diverse reconstructions},
  author    = {Morshuis, Jan Nikolas and Hein, Matthias and Baumgartner, Christian F},
  booktitle = {Proc. {MICCAI} Workshop Deep Gen. Models},
  pages     = {180--190},
  year      = {2024}
}

@article{huang_considering_2021,
  title   = {Considering anatomical prior information for low-dose {CT} image enhancement using attribute-augmented {Wasserstein} generative adversarial networks},
  volume  = {428},
  journal = {Neurocomputing},
  author  = {Huang, Zhenxing and Liu, Xinfeng and Wang, Rongpin and Chen, Jincai and Lu, Ping and Zhang, Qiyang and Jiang, Changhui and Yang, Yongfeng and Liu, Xin and Zheng, Hairong and Liang, Dong and Hu, Zhanli},
  year    = {2021},
  pages   = {104--115}
}

@article{ongie_deep_2020,
  title   = {Deep learning techniques for inverse problems in imaging},
  volume  = {1},
  number  = {1},
  journal = {IEEE J. Sel. Areas Inf. Theory},
  author  = {Ongie, Gregory and Jalal, Ajil and Metzler, Christopher A. and Baraniuk, Richard G. and Dimakis, Alexandros G. and Willett, Rebecca},
  year    = {2020},
  pages   = {39--56}
}

@article{antun_instabilities_2020,
  title   = {On instabilities of deep learning in image reconstruction and the potential costs of {AI}},
  volume  = {117},
  number  = {48},
  journal = {Proc. Nat. Acad. Sci.},
  author  = {Antun, Vegard and Renna, Francesco and Poon, Clarice and Adcock, Ben and Hansen, Anders C.},
  year    = {2020},
  pages   = {30088--30095}
}

@article{suomalainen_dentomaxillofacial_2015,
  title   = {Dentomaxillofacial imaging with panoramic views and cone beam {CT}},
  volume  = {6},
  number  = {1},
  journal = {Insights Imaging},
  author  = {Suomalainen, Anni and Pakbaznejad Esmaeili, Elmira and Robinson, Soraya},
  year    = {2015},
  pages   = {1--16}
}

@article{scarfe_what_2008,
  title   = {What is {Cone}-{Beam} {CT} and {How} {Does} it {Work}?},
  volume  = {52},
  number  = {4},
  journal = {Dent. Clin. North Am.},
  author  = {Scarfe, William C. and Farman, Allan G.},
  year    = {2008},
  pages   = {707--730}
}

@inproceedings{thongvigitmanee_radiation_2013,
  address   = {Osaka},
  title     = {Radiation dose and accuracy analysis of newly developed cone-beam {CT} for dental and maxillofacial imaging},
  booktitle = {Proc. 35th Annu. Int. Conf. {IEEE} Eng. Med. Biol. Soc. ({EMBC})},
  author    = {Thongvigitmanee, Saowapak S. and Pongnapang, Napapong and Aootaphao, Sorapong and Yampri, Pinyo and Srivongsa, Tanapong and Sirisalee, Pasu and Rajruangrabin, Jartuwat and Thajchayapong, Pairash},
  year      = {2013},
  pages     = {2356--2359}
}

@article{leung_accuracy_2010,
  title   = {Accuracy and reliability of cone-beam computed tomography for measuring alveolar bone height and detecting bony dehiscences and fenestrations},
  volume  = {137},
  number  = {4},
  journal = {Am. J. Orthod. Dentofacial Orthop.},
  author  = {Leung, Cynthia C. and Palomo, Leena and Griffith, Richard and Hans, Mark G.},
  year    = {2010},
  pages   = {S109--S119}
}

@article{izzetti_basic_2021,
  title   = {Basic {Knowledge} and {New} {Advances} in {Panoramic} {Radiography} {Imaging} {Techniques}: {A} {Narrative} {Review} on {What} {Dentists} and {Radiologists} {Should} {Know}},
  volume  = {11},
  number  = {17},
  journal = {Appl. Sci.},
  author  = {Izzetti, Rossana and Nisi, Marco and Aringhieri, Giacomo and Crocetti, Laura and Graziani, Filippo and Nardi, Cosimo},
  year    = {2021},
  pages   = {7858}
}

@article{ludlow_patient_2008,
  address = {1939},
  title   = {Patient risk related to common dental radiographic examinations: the impact of 2007 {International} {Commission} on {Radiological} {Protection} recommendations regarding dose calculation},
  number  = {9},
  journal = {J. Am. Dent. Assoc.},
  author  = {Ludlow, John B. and Davies-Ludlow, Laura E. and White, Stuart C.},
  year    = {2008},
  pages   = {1237--1243}
}

@article{mohamed_3d_2025,
  title   = {{3D} reconstruction from {2D} multi-view dental images based on {EfficientNetB0} model},
  volume  = {15},
  number  = {1},
  journal = {Sci. Rep.},
  author  = {Mohamed, Waleed and Nader, Nermeen and Alsakar, Yasmin M. and Elazab, Naira and Ezzat, Mohamed and Elmogy, Mohammed},
  year    = {2025},
  pages   = {28775}
}

@article{lopes_comparison_nodate,
  title     = {Comparison of panoramic radiography and {CBCT} to identify maxillary posterior roots invading the maxillary sinus},
  author    = {Lopes, Luciana J and Gamba, Thiago O and Bertinato, Joao VJ and Freitas, Deborah Q},
  journal   = {Dentomaxillofac. Radiol.},
  volume    = {45},
  number    = {6},
  pages     = {20160043},
  year      = {2016},
  publisher = {Oxford University Press}
}

@article{tandogdu_comparison_2022,
  title   = {Comparison of the {Efficacy} of the {Panoramic} and {Cone} {Beam} {Computed} {Tomography} {Imaging} {Methods} in the {Surgical} {Planning} of the {Maxillary} {All}-{On}-4, {M}-4, and {V}-4},
  volume  = {2022},
  journal = {BioMed Res. Int.},
  author  = {Tandogdu, Erim and Ayali, Aysa and Caymaz, Mehmet Gagari},
  year    = {2022},
  pages   = {1553340}
}

@article{antony_two-dimensional_nodate,
  title   = {Two-dimensional {Periapical}, {Panoramic} {Radiography} {Versus} {Three}-dimensional {Cone}-beam {Computed} {Tomography} in the {Detection} of {Periapical} {Lesion} {After} {Endodontic} {Treatment}: {A} {Systematic} {Review}},
  volume  = {12},
  number  = {4},
  journal = {Cureus},
  author  = {Antony, Delphine P and Thomas, Toby and Nivedhitha, MS},
  year    = {2020},
  pages   = {e7736}
}

@inproceedings{hatamizadeh_unetr_2022,
  title     = {{UNETR}: {Transformers} for {3D} medical image segmentation},
  booktitle = {Proc. {IEEE}/{CVF} Winter Conf. Appl. Comput. Vis. ({WACV})},
  author    = {Hatamizadeh, Ali and Tang, Yucheng and Nath, Vishwesh and Yang, Dong and Myronenko, Andriy and Landman, Bennett and Roth, Holger R. and Xu, Daguang},
  year      = {2022},
  pages     = {574--584}
}

@inproceedings{linguraru_3dpx_2024,
  title     = {{3DPX}: {Progressive} {2D}-to-{3D} {Oral} {Image} {Reconstruction} with {Hybrid} {MLP}-{CNN} {Networks}},
  volume    = {15007},
  booktitle = {Proc. Med. Image Comput. Comput. Assist. Intervent. ({MICCAI})},
  author    = {Li, Xiaoshuang and Meng, Mingyuan and Huang, Zimo and Bi, Lei and Delamare, Eduardo and Feng, Dagan and Sheng, Bin and Kim, Jinman},
  year      = {2024},
  pages     = {25--34},
  publisher = {Springer}
}

@inproceedings{wang_semit-sam_2025,
  title     = {{SemiT}-{SAM}: {Building} {A} {Visual} {Foundation} {Model} for {Tooth} {Instance} {Segmentation} on {Panoramic} {Radiographs}},
  volume    = {15571},
  booktitle = {Proc. Supervised Semi-Supervised Multi-Struct. Segment. Landmark Detect. Dental Data},
  author    = {Hao, Jing and Liu, Moyun and He, Lei and Yao, Lei and Tsoi, James Kit Hon and Hung, Kuo Feng},
  year      = {2025},
  pages     = {110--121},
  publisher = {Springer}
}

@article{ma_segment_2024,
  title   = {Segment anything in medical images},
  volume  = {15},
  number  = {1},
  journal = {Nat. Commun.},
  author  = {Ma, Jun and He, Yuting and Li, Feifei and Han, Lin and You, Chenyu and Wang, Bo},
  year    = {2024},
  pages   = {654}
}

@inproceedings{kirillov_segment_2023,
  title     = {Segment anything},
  booktitle = {Proc. {IEEE}/{CVF} Int. Conf. Comput. Vis. ({ICCV})},
  author    = {Kirillov, Alexander and Mintun, Eric and Ravi, Nikhila and Mao, Hanzi and Rolland, Chloe and Gustafson, Laura and Xiao, Tete and Whitehead, Spencer and Berg, Alexander C. and Lo, Wan-Yen},
  year      = {2023},
  pages     = {4015--4026}
}

@inproceedings{avidan_conditional-flow_2022,
  title     = {Conditional-{Flow} {NeRF}: {Accurate} {3D} {Modelling} with {Reliable} {Uncertainty} {Quantification}},
  volume    = {13663},
  booktitle = {Proc. Comput. Vis. -- {ECCV}},
  author    = {Shen, Jianxiong and Agudo, Antonio and Moreno-Noguer, Francesc and Ruiz, Adria},
  year      = {2022},
  pages     = {540--557},
  publisher = {Springer}
}

@inproceedings{ying_x2ct-gan_2019,
  title     = {{X2CT}-{GAN}: reconstructing {CT} from biplanar {X}-rays with generative adversarial networks},
  booktitle = {Proc. {IEEE}/{CVF} Conf. Comput. Vis. Pattern Recognit. ({CVPR})},
  author    = {Ying, Xingde and Guo, Heng and Ma, Kai and Wu, Jian and Weng, Zhengxin and Zheng, Yefeng},
  year      = {2019},
  pages     = {10619--10628}
}

@inproceedings{cipriano_improving_2022,
  title     = {Improving segmentation of the inferior alveolar nerve through deep label propagation},
  booktitle = {Proc. {IEEE}/{CVF} Conf. Comput. Vis. Pattern Recognit. ({CVPR})},
  author    = {Cipriano, Marco and Allegretti, Stefano and Bolelli, Federico and Pollastri, Federico and Grana, Costantino},
  year      = {2022},
  pages     = {21137--21146}
}

@article{lumetti_enhancing_2024,
  title   = {Enhancing patch-based learning for the segmentation of the mandibular canal},
  volume  = {12},
  journal = {IEEE Access},
  author  = {Lumetti, Luca and Pipoli, Vittorio and Bolelli, Federico and Ficarra, Elisa and Grana, Costantino},
  year    = {2024},
  pages   = {79014--79024}
}

@article{bolelli_segmenting_2024,
  title   = {Segmenting the inferior alveolar canal in {CBCTs} volumes: the toothfairy challenge},
  author  = {Bolelli, Federico and Lumetti, Luca and Vinayahalingam, Shankeeth and Di Bartolomeo, Mattia and Pellacani, Arrigo and Marchesini, Kevin and Van Nistelrooij, Niels and Van Lierop, Pieter and Xi, Tong and Liu, Yusheng and others},
  journal = {IEEE Trans. Med. Imaging},
  volume  = {44},
  number  = {4},
  pages   = {1890--1906},
  year    = {2024}
}

@article{wang_mmdental-multimodal_2025,
  title   = {{MMDental}-{A} multimodal dataset of tooth {CBCT} images with expert medical records},
  volume  = {12},
  number  = {1},
  journal = {Sci. Data},
  author  = {Wang, Chengkai and Zhang, Yifan and Wu, Chengyu and Liu, Jun and Huang, Xingliang and Wu, Liuxi and Wang, Yitong and Feng, Xiang and Lu, Yiting and Wang, Yaqi},
  year    = {2025},
  pages   = {1172}
}

@article{cui_fully_2022,
  title   = {A fully automatic {AI} system for tooth and alveolar bone segmentation from cone-beam {CT} images},
  volume  = {13},
  number  = {1},
  journal = {Nat. Commun.},
  author  = {Cui, Zhiming and Fang, Yu and Mei, Lanzhuju and Zhang, Bojun and Yu, Bo and Liu, Jiameng and Jiang, Caiwen and Sun, Yuhang and Ma, Lei and Huang, Jiawei},
  year    = {2022},
  pages   = {2096}
}

@article{scarfe_maxillofacial_2012,
  title   = {Maxillofacial cone beam computed tomography: essence, elements and steps to interpretation},
  volume  = {57},
  number  = {s1},
  journal = {Aust. Dent. J.},
  author  = {Scarfe, Wc and Li, Z and Aboelmaaty, W and Scott, Sa and Farman, Ag},
  year    = {2012},
  pages   = {46--60}
}

@inproceedings{corona-figueroa_mednerf_2022,
  author    = {Corona-Figueroa, Abril and Frawley, Jonathan and Taylor, Sam Bond- and Bethapudi, Sarath and Shum, Hubert P. H. and Willcocks, Chris G.},
  booktitle = {Proc. 44th Annu. Int. Conf. {IEEE} Eng. Med. Biol. Soc. ({EMBC})},
  title     = {{MedNeRF}: Medical Neural Radiance Fields for Reconstructing {3D}-aware {CT}-Projections from a Single {X-ray}},
  year      = {2022},
  pages     = {3843-3848}
}

@article{balakrishnan_voxelmorph_2019,
  title   = {{Voxelmorph}: a learning framework for deformable medical image registration},
  volume  = {38},
  number  = {8},
  journal = {IEEE Trans. Med. Imaging},
  author  = {Balakrishnan, Guha and Zhao, Amy and Sabuncu, Mert R. and Guttag, John and Dalca, Adrian V.},
  year    = {2019},
  pages   = {1788--1800}
}

@book{mallya_white_2019,
  title     = {White and {Pharoah}'s Oral Radiology: Principles and Interpretation},
  author    = {Mallya, Sanjay M. and Lam, Ernest W. N.},
  edition   = {8th},
  year      = {2019},
  publisher = {Elsevier},
  address   = {St. Louis, MO, USA}
}

@article{jaju_cone-beam_2015,
  title   = {Cone-beam computed tomography: {Time} to move from {ALARA} to {ALADA}},
  volume  = {45},
  number  = {4},
  journal = {Imaging Sci. Dent.},
  author  = {Jaju, Prashant P. and Jaju, Sushma P.},
  year    = {2015},
  pages   = {263}
}

@inproceedings{linguraru_px2tooth_2024,
  title     = {{PX2Tooth}: {Reconstructing} the {3D} {Point} {Cloud} {Teeth} from a {Single} {Panoramic} {X}-{Ray}},
  volume    = {15003},
  author    = {Ma, Wen and Wu, Huikai and Xiao, Zikai and Feng, Yang and Wu, Jian and Liu, Zuozhu},
  booktitle = {Proc. Med. Image Comput. Comput. Assist. Intervent. ({MICCAI})},
  year      = {2024},
  pages     = {411--421},
  publisher = {Springer}
}

@inproceedings{song_oral-3d_2021,
  title     = {Oral-{3D}: {Reconstructing} the {3D} structure of oral cavity from panoramic x-ray},
  volume    = {35},
  booktitle = {Proc. {AAAI} Conf. Artif. Intell.},
  author    = {Song, Weinan and Liang, Yuan and Yang, Jiawei and Wang, Kun and He, Lei},
  year      = {2021},
  pages     = {566--573}
}

@inproceedings{martel_x2teeth_2020,
  title     = {{X2Teeth}: {3D} {Teeth} {Reconstruction} from a {Single} {Panoramic} {Radiograph}},
  volume    = {12262},
  booktitle = {Proc. Med. Image Comput. Comput. Assist. Intervent. ({MICCAI})},
  author    = {Liang, Yuan and Song, Weinan and Yang, Jiawei and Qiu, Liang and Wang, Kun and He, Lei},
  year      = {2020},
  pages     = {400--409},
  publisher = {Springer}
}

@article{parida_vit-nebla_2025,
  author  = {Keshari Parida, Bikram and Sunilkumar, Anusree P. and Sen, Abhijit and You, Wonsang},
  journal = {IEEE Access},
  title   = {{ViT-NeBLa}: A Hybrid Vision Transformer and Neural Beer–Lambert Framework for Single-View {3D} Reconstruction of Oral Anatomy From Panoramic Radiographs},
  year    = {2025},
  volume  = {13},
  pages   = {170761-170781}
}

@inproceedings{park_3d_2023,
  title     = {{3D} {Teeth} {Reconstruction} from {Panoramic} {Radiographs} {Using} {Neural} {Implicit} {Functions}},
  booktitle = {Lect. Notes Comput. Sci.},
  author    = {Park, Sihwa and Kim, Seongjun and Song, In-Seok and Baek, Seung Jun},
  year      = {2023},
  volume    = {14229},
  pages     = {376--386},
  publisher = {Springer}
}

@article{mei_dtr-net_2024,
  title   = {{DTR}-{Net}: {Dual}-{Space} {3D} {Tooth} {Model} {Reconstruction} {From} {Panoramic} {X}-{Ray} {Images}},
  volume  = {43},
  number  = {1},
  journal = {IEEE Trans. Med. Imaging},
  author  = {Mei, Lanzhuju and Fang, Yu and Zhao, Yue and Zhou, Xiang Sean and Zhu, Min and Cui, Zhiming and Shen, Dinggang},
  year    = {2024},
  pages   = {517--528}
}
\end{document}